\documentclass[letterpaper, 10pt, conference]{ieeeconf}      

\IEEEoverridecommandlockouts                              
\usepackage{times} 
\usepackage{graphicx} 
\usepackage{bm} 
\usepackage{amsmath} 
\usepackage{amssymb} 
\usepackage{color} 
\usepackage[ruled,linesnumbered]{algorithm2e} 
\usepackage[short]{optidef} 
\usepackage{gensymb} 
\usepackage{multirow}
\usepackage[mathscr]{eucal}

\usepackage{xparse,soul} 
\soulregister\cite7 
\soulregister\citep7 
\soulregister\citet7 
\soulregister\ref7 
\soulregister\pageref7 

\usepackage[flushleft]{threeparttable}
\usepackage{makecell}

\usepackage{fancyhdr}
\fancypagestyle{firstpage}{
    \chead{This paper has been accepted for publication at the 2025 International Conference on Intelligent Robots and Systems.}}
\usepackage{cite}

\title{Augmented Bridge Spinal Fixation: A New Concept for Addressing Pedicle Screw Pullout via a Steerable Drilling Robot and  Flexible Pedicle Screws}
\author{Yash Kulkarni$^{1}$, Susheela Sharma$^{1}$, Omid Rezayof$^{1}$, Siddhartha Kapuria$^{1}$, Jordan P. Amadio$^{2}$, \\ Mohsen Khadem$^{3}$, Maryam Tilton$^{1}$, and Farshid Alambeigi$^{1}$ \IEEEmembership{Member, IEEE}
\thanks{*This work was supported in part by the National Institute Of Biomedical Imaging and Bioengineering of the National Institutes of Health under Award Number R21EB030796 and in part by the Collaborative Accelerator for Transformative Research Endeavors grant, jointly awarded by The University of Texas at Austin and The University of Texas MD Anderson Cancer Center.}
\thanks{$^{1}$Y.~Kulkarni, S.~Sharma, O.~Rezayof, S.~Kapuria, M.~Tilton and F.~Alambeigi are with the Walker Department of Mechanical Engineering at the University of Texas at Austin, Austin, TX, 78712, USA. Email: \{kulkarni.yash08, sheela.sharma, omid.rezayof, skapuri\}@utexas.edu,\{maryam.tilton, farshid.alambeigi\}@austin.utexas.edu}.
\thanks{$^{2}$J.~P.~ Amadio is with the Department of Neurosurgery, The University of Texas Dell Medical School, TX, 78712.}
\thanks{$^{3}$M.~Khadem is with the School of Informatics, University of Edinburgh, UK. }}

\usepackage{color} 

\begin{document}
\maketitle
\thispagestyle{firstpage}
\pagestyle{empty}
		
\begin{abstract}
To address the  screw loosening and pullout limitations  of rigid pedicle screws in spinal fixation procedures, and to leverage our recently developed  Concentric Tube Steerable Drilling Robot (CT-SDR) and Flexible Pedicle Screw (FPS), in this paper, we introduce the concept of  Augmented Bridge Spinal Fixation (AB-SF). In this concept,  two connecting J-shape tunnels are first drilled through pedicles of vertebra using the CT-SDR. Next, two FPSs are passed through this tunnel and bone cement is then injected through the cannulated region of the FPS to form an augmented bridge between two pedicles and reinforce strength of the fixated spine. To experimentally analyze and study the feasibility of AB-SF technique, we first used our robotic system (i.e., a CT-SDR integrated with a robotic arm) to create two different fixation scenarios in which two J-shape tunnels, forming a bridge, were drilled at different depth of a vertebral phantom. Next, we implanted two FPSs within the drilled tunnels and then successfully simulated the bone cement augmentation process. 
 
\end{abstract}

\section{Introduction}
\begin{figure}[t] 
    \centering 
    \includegraphics[width=1\linewidth]{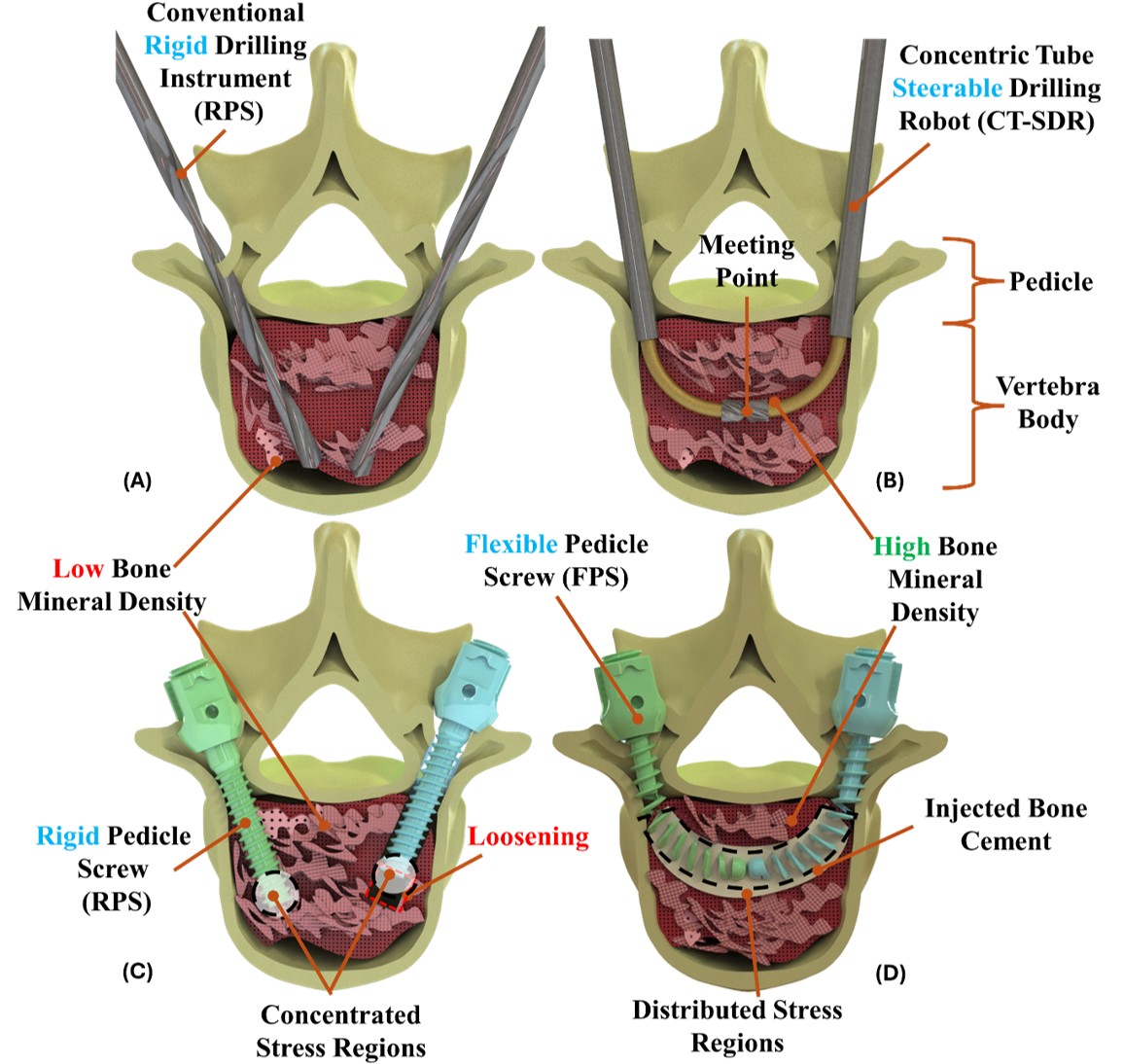}
    \caption{Conceptual illustration of the proposed Augmented Bridge Spinal Fixation (AB-SF) approach: (a) and (b) compare the existing straight drilling using rigid instruments with the bridge drilling concept using CT-SDR, respectively.  Also, figures (c) and (d) compare the existing RPS fixation with the proposed AB-SF concept using bone cement.}
    \label{fig:concept}
\end{figure}

In people over the age of 50, vertebral compression fractures are the most prevalent type of osteoporotic-related bone fracture (i.e., 1.4 million global occurrences)  \cite{johnell2006estimate}. The standard treatment for vertebral compression fractures is spinal fixation (SF) surgery. SF  is a multi-step procedure that enables two or more vertebral bodies to fuse together in order to return stability back to the affected patient. 
As illustrated in Fig. \ref{fig:concept}-A and Fig. \ref{fig:concept}-C, this procedure involves drilling two separate, straight holes within the pedicle canals of the vertebra, using rigid drilling instruments.
Next, two Rigid Pedicle Screws (RPSs) are carefully inserted through the drilled, non-connected tunnels and fixated into the cancellous bone regions of vertebra. Success of SF procedures directly depends on the accuracy of the drilling procedure and bone quality in the vertebral body defined by its bone mineral density (BMD) \cite{Sharma2024TBME}.   While SF surgery using RPSs has become the gold-standard for fixating vertebral compression fractures, RPS \textit{loosening and pullout} still remain critical problems of current SF procedures in osteoporotic patients suffering from low BMD \cite{weiser2017insufficient}.  

Researchers have explored various methods to prevent the RPS loosening and pullout problem ranging from changing the RPS design parameters \cite{Talu2000PedicleSS, Mehta2012BiomechanicalAO}
to using computer assisted methods to plan the insertion trajectories of screws within the vertebral body \cite{Knez2016ComputerAssistedSS}. However, these approaches have largely been ineffective. While pedicle screw augmentation with bone cement (PMMA) has demonstrated greater success in reducing pedicle screw pullout and loosening by providing additional anchorage \cite{Kim2020ClinicalEA}, it does not fully resolve the issue as it only reinforces the immediate area around the screw. This leaves the overall fixation dependent on a small region of augmented bone. In osteoporotic patients, where bone quality is already compromised, this localized reinforcement may still be insufficient to maintain long-term stability. Consequently, as reported by Wieser et al. \cite{weiser2017insufficient}, over 90\% of SF procedures in osteoporotic vertebrae ultimately fail, necessitating revision surgery.

One of the main reasons for pedicle screw loosening and pullout lies in the anatomical constraints of vertebrae and rigidity of existing RPSs and  instruments \cite{Sharma2023TBME}. Such limitations constrain clinicians to fixate these rigid implants in a limited space within the vertebral body that may be osteoporotic or have low BMD regions \cite{Alambeigi2018InroadsTR,alambeigi2019use,Sharma2023ISMR,Sharma2023TBME}. Aside from the rigidity of used instruments, another reason for such failures can be attributed to the nature of stress handling in the existing \textit{disconnected} RPS fixation methods. In a fixated spine, loading on the spine is transferred to the implanted RPSs and stress concentrations happen at the discrete regions around the tips of the RPSs located inside the vertebral body (see Fig. \ref{fig:concept}-C). The cancellous bone around the screw tips must handle the \textit{locally concentrated maximum stresses} at these regions. Otherwise, the screw starts to loosen at these locations, which typically occurs in osteoporotic patients \cite{Law1993CaudocephaladLO}.

To address the access limitation of existing drilling instruments, researchers have started developing different types of steerable drilling systems. For instance, Wang et al. \cite{Hinged2} introduced an articulated hinged drilling tool to increase access in the vertebral body, but is unable to create smooth curved trajectories with this method. Alambeigi et al. \cite{alambeigi2017curved,alambeigi2019use} also introduced a tendon driven steerable drilling robot (SDR) capable of creating smooth trajectories. Nevertheless, this system lacks sufficient structural strength and is not intuitive to be used by clinicians. To address such limitations, we have recently designed and successfully evaluated many Concentric Tube Steerable Drilling Robot (CT-SDR) \cite{Sharma2023TBME,Sharma2023ICRA,Sharma2023ISMR,Sharma2024TBME, Maroufi2025S3D,Kulkarni2025TowardsDD}. Thanks to its  design features, the CT-SDR is intuitive to use and provides simultaneous structural stiffness and compliance. It also provides a smooth drilling trajectory for creating  curved J- shape tunnels for implanting a flexible implant within the drilled curved trajectory.  
Leveraging the capabilities of the CT-SDR, we have also recently introduced a unique complementary Flexible Pedicle Screw (FPS)  that can be implanted inside a J-shape drilled trajectory made by the CT-SDR\cite{Kulkarni2025SynergisticPSA,Kulkarni2024TowardsBE,yash2024SFF}. As shown in Fig. \ref{fig:concept}-D, FPS are biocompatible, semi-rigid, semi-flexible implants that can be additively manufactured using titanium or stainless steel. It also can morph within a J-shape drilled tunnel to successfully distribute the load along the spine's geometry -- as opposed to the mentioned discrete stress handling of existing RPSs.  

To address the screw loosening and pullout of RPSs and leveraging the features of our CT-SDR and FPS, in this paper and as our main contribution, we introduce the concept of  \textit{Augmented Bridge Spinal Fixation (AB-SF)}. As shown in Fig. \ref{fig:concept}-B and Fig. \ref{fig:concept}-D, in this approach, first two connecting J-shape tunnels are drilled through the pedicles of the vertebra using the CT-SDR. Next, two FPSs are passed through this tunnel and bone cement is  injected through the center of the FPS to create an augmented bridge between the screws. The bone cement augmentation provides a reinforced internal connection between two pedicles and can potentially increase the fixation strength and minimize the risk of screw loosening and pullout failures by distributing the stress along the augmented bridge. To experimentally analyze and study the feasibility of AB-SF technique, we first used our robotic system (i.e., a CT-SDR integrated with a robotic arm) to create two different fixation scenarios in which two J-shape tunnels, forming a bridge, were drilled at different depth of a vertebral phantom. Next, we implanted two FPSs within the drilled tunnels and then successfully simulated the bone cement augmentation process.

\begin{figure}[t] 
    \centering 
    \includegraphics[width=1\linewidth]{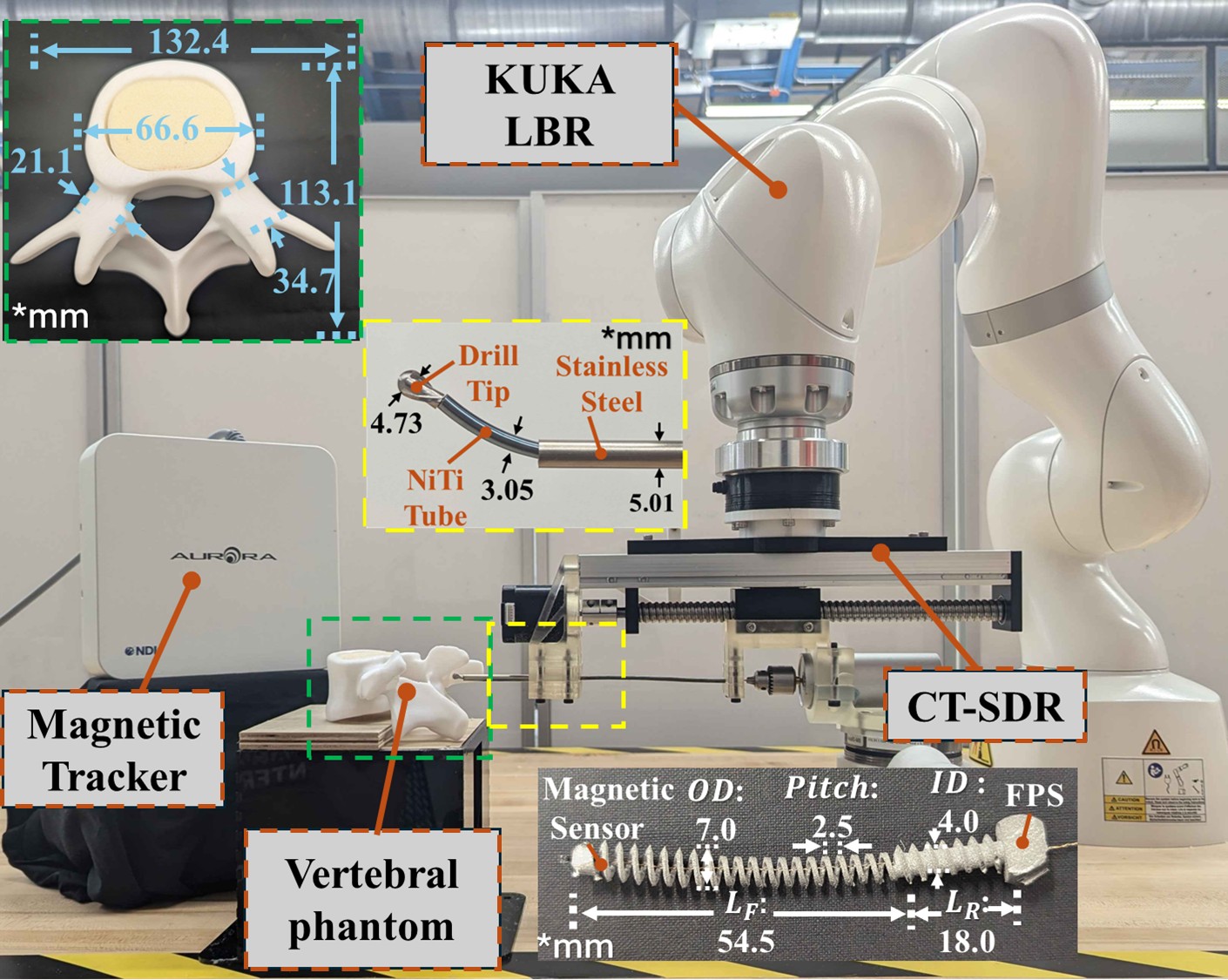}
    \caption{Experimental set-up used for evaluating the proposed AB-SF concept, including a CT-SDR, a robotic manipulator (KUKA LBR Med), an FPS, and Aurora Magnetic Tracker, and a custom-designed vertebral phantom. All units are in mm.}
    \label{fig:Experiment}
\end{figure}

\section{Steerable Drilling Robotic System}
Fig. \ref{fig:Experiment} shows the CT-SDR robotic system  used in this study. The system consists of the  CT-SDR \cite{Sharma2023TBME} attached to a 7 Degree of Freedom (DoF) KUKA LBR Med 7 (KUKA, Germany) system. 
This combination is vital in creating trajectories with both a straight and curved component (i.e., J-shape trajectory) to match the design of the FPS shown in Fig. \ref{fig:Experiment} for the proposed AB-SF concept.   
As shown in Fig. \ref{fig:Experiment}, the CT-SDR consists of (i) a pre-shaped Nitinol (NiTi) tube constrained inside a straight stainless steel tube, which defines the drilling trajectory, (ii) a flexible drilling tool to create the specified curved trajectories, (iii) a linear stage that controls the insertion and retraction of the NiTi tube, and (iv) a drill motor controlling the rotational speed of the flexible drilling tool. A 3D printed coupling (PLA, Raise3D) is used to connect the CT-SDR to the robotic manipulator. 
The desired shape of the heat treated NiTi tube is heavily affected by the geometry of the planned AB-SF trajectory (described in Section \ref{section:Trajectory}). For this study and without loss of generality, a NiTi tube with a 3.05 mm diameter and a 0.24 mm wall thickness was used. Once the NiTi tube is properly heat treated and fixed within the CT-SDR system, the flexible cutting tool is safely fed through and constrained within the NiTi tube. The flexible cutting tools consists of a 4.73 mm outer diameter  ball nose end mill (McMaster-Carr), laser welded to a flexible torque coil (Asahi Intec. USA, Inc.). These values are shown on Fig. \ref{fig:Experiment}.

While the actuation unit for the individual CT-SDR allows for 1 DoF insertion and retraction movement, the robotic arm provides additional DoFs to place the CT-SDR in the workspace and create the J-shape drilling trajectories necessary for the proposed AB-SF concept.   Driven by the design of the FPS, which requires two different trajectories (straight and curved) to be drilled within the vertebral body, multiple control strategies were implemented with the CT-SDR system. First, to  create a straight trajectory, an \textit{Admittance Control} strategy was implemented. This strategy enables the robotic arm to move freely through the workspace based on the surgeons inputs. This enables the surgeon to position the robot in their desired alignment. 
Next, after the robot's alignment is finalized, the \textit{Autonomous Drilling Control} strategy is employed. Here, the manipulator autonomously drills a straight trajectory corresponding to the straight length chosen for the AB-SF. Finally, after the straight trajectory is drilled, the \textit{Stationary Control} strategy is employed. In this phase, the manipulator holds its position while the CT-SDR drills into the vertebra to create the curved portion of the desired trajectory chosen for the AB-SF. 

\section{Trajectory Planning in AB-SF Technique} \label{section:Trajectory}
As shown in Fig. \ref{fig:concept}-A and Fig. \ref{fig:concept}-C, in a routine SF surgery, the planning of RPS placement is constrained to a linear insertion region that is both safe and accessible with traditional rigid drilling instruments. However, our previously introduced CT-SDR  \cite{Sharma2023TBME,Sharma2023ISMR} has demonstrated the ability to safely deviate from these rigid linear path to fixate an FPS inside high BMD areas of vertebra. This flexibility enables the creation of non-linear and curved trajectories, such as the envisioned AB-SF configuration shown in Fig. \ref{fig:concept}-B and Fig. \ref{fig:concept}-D. However, unlike our previous approaches \cite{Sharma2024SpatialSF}, the key challenge in this new approach is ensuring that two flexible pedicle screws meet accurately within the vertebral body at the desired meeting point. Given the need for both screws to meet in a precise configuration, careful trajectory planning is critical. As shown in Fig. \ref{fig:Param}, effective planning ensures that both the CT-SDR and the FPS are aligned properly so that the screws meet precisely in the AB-SF configuration. In this section, we will provide more details for the parameter selection and procedure planning process to ensure a successful AB-SF Procedure. 

\subsection{Adjustable Design Parameters and Constraints}
\begin{figure}[t] 
    \centering 
    \includegraphics[width=1\linewidth]{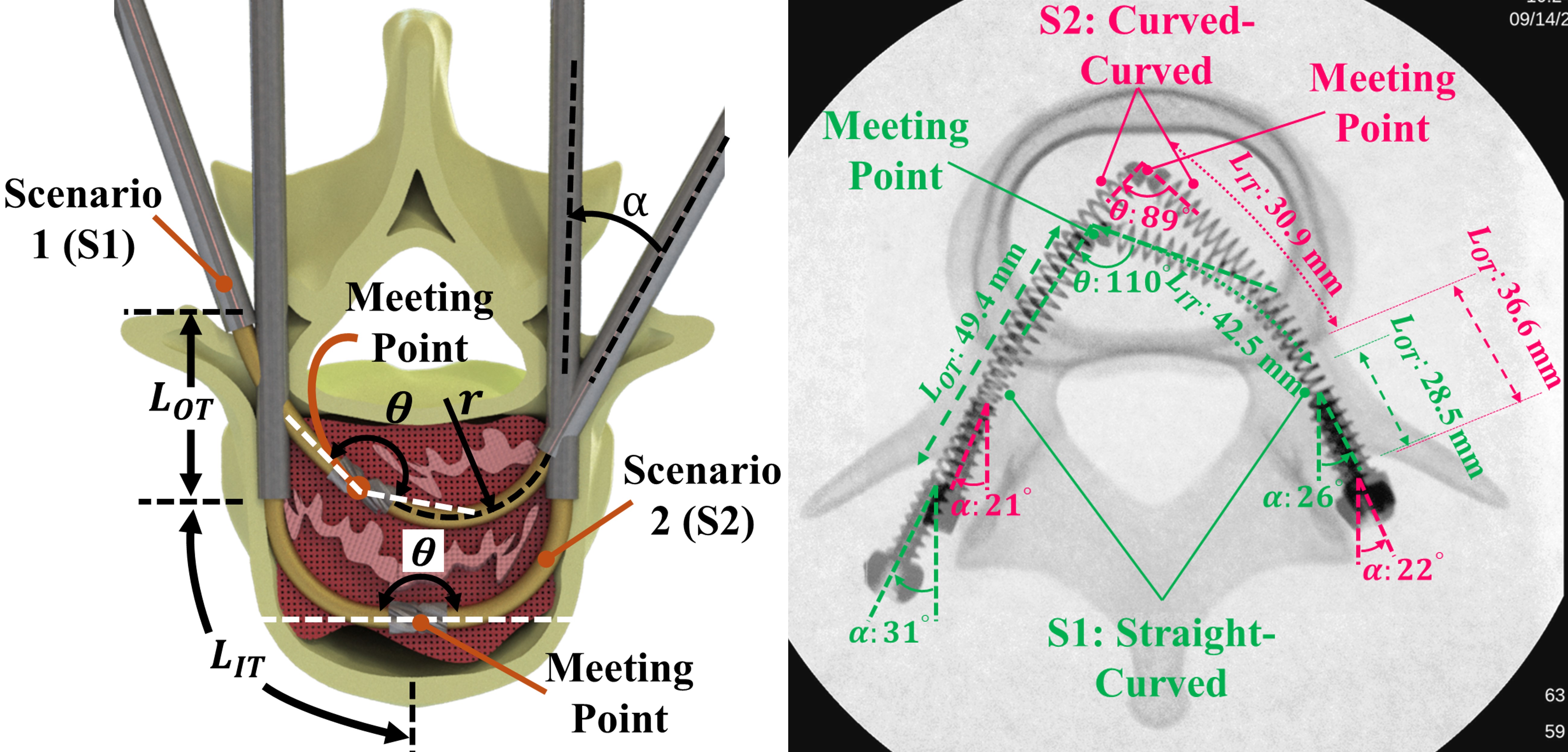}
    \caption{ (Left) The design parameters required for planning a AB-SF drilling trajectory. (Right) X-ray overlay of the executed AB-SF scenarios inside a Sawbone phantom comparing their trajectories and insertion depth.}
    \label{fig:Param}
\end{figure}

To create the necessary connecting curved trajectories for the AB-SF concept, the CT-SDR system is governed by five adjustable parameters and four key constraints. These parameters and constraints play a pivotal role in ensuring that the CT-SDR can successfully navigate inside the complex anatomy of vertebra and form the envisioned trajectories for a successful AB-SF procedure. 
The \textit{first} adjustable parameter is the insertion angle of the CT-SDR along the axial plane ($\alpha$). As shown on Fig. \ref{fig:Param}, $\alpha$ is the angle spanning the entire pedicle region as the entrance canal of the  CT-SDR into vertebral body. As illustrated by the two distinct scenarios on Fig. \ref{fig:Param}, the $\alpha$ plays a critical role in determining the initial approaching angle of CT-SDR into the vertebral body. A small and large $\alpha$ directly affects and define the trajectory  of the CT-SDR and allows the surgeon to target specific regions of the vertebral body.  The \textit{second} and \textit{third} adjustable parameters are the insertion depth of the outer tube ($L_{OT}$) and inner tube of the CT-SDR ($L_{IT}$). $L_{OT}$ plays a critical role in controlling depth of the straight section of a J-shape trajectory created by the CT-SDR while $L_{IT}$ plays a vital role in controlling the final depth and location of the curved trajectory. The \textit{fourth} adjustable parameter is the  radius of curvature for the inner tube ($r$). This parameter controls the sharpness of the  created curvature. A smaller $r$ allows the creation of a more sharper turn thus allowing us to access more sharper corners of the vertebral body. The final adjustable parameter is the meeting angle of the drill bits  ($\theta$). This  is defined based on the angles formed between the  tangent lines of the created tunnel from both side of pedicles. 

In other words, this angle dictates how the drill bits and consequently two FPSs can meet and still form an acceptable AB-SF formation.   

Moreover, there are four constraints that guide the selection of the aforementioned adjustable parameters. Two of these constraints are patient specific: the anatomy of the vertebra and the osteoporotic regions within the vertebral body. As each patient has a unique anatomy and distribution of osteoporosis, this constraints have a direct affect on both the CT-SDR and the FPS design. The vertebral anatomy and level determine angle $\alpha$, maximum insertion length and radius of curvature for the curved section of the J-shape trajectories, and diameter of the drilled tunnel.  The osteoporotic regions are also critical to consider since drilling and fixating in this weakened regions can lead to poor fixation. Locations of osteoporotic regions directly affect the the maximum insertion length and radius of curvature of the drilled tunnel. The remaining two constraints are system specific: the limitations on the smallest possible radius of curvatures the CT-SDR system can create and the meeting angles at which the CT-SDRs can meet during the drilling. Due to the manufacturing constraints of the inner tube, only certain curvatures are possible to create thus restricting the range of angles the system is capable of drilling.

\subsection{Planning Procedure}
Considering our use case, there are five critical steps in ensuring successful creation of our envisioned AB-SF spinal fixation structure. First (i) the surgeon must analyze the patient's vertebra with quantitative CT scans to establish areas of high BMD and locate the optimal area for pedicle screw placement. This is vital in ensuring osteoporotic regions of the vertebra body are avoided and stable fixation can occur. Then (ii) based on the anatomy of the vertebra body and the geometrical constraints of the pedicle, the proper $\alpha$ value must be chosen to avoid any undesirable interaction with the pedicle wall and spinal nerves. After the proper insertion angle is finalized, (iii) the FPS placement must be choosen such that they avoid any undesirable location and reach the optimal location for fixation. This can be achieved by modifying $L_{OT}$, $L_{IT}$, and $r$ as required, while ensuring the $\theta$ is within bounds. Once the optimal trajectory is finalized, (iv) our previously introduced \textit{Biomechanics-Aware Trajectory Selection Module} in our \textit{Biomechanics-Aware Robot-Assisted Drilling} Framework \cite{Sharma2024TBME} can be used to ascertain the stress and strain distribution along the selected trajectory. This will help ensure the selected path presents the best location for SF for the individual patient. Finally, (v) the corresponding FPS can be selected to ensure it matches the curved trajectory length and the CT-SDR can be assembled to the parameters selected to create the designed trajectory.

\section{Experimental Set-up and Results}
To verify that the FPS could reliably follow the planned AB-SF trajectory, we considered  2 distinct drilling scenarios. These scenarios are vital to evaluate our steerable drilling robotic system's ability to create the designed connecting trajectories and the  FPS ability to conform to this drilled path. These drilling scenarios demonstrate different potential AB-SF trajectories with two distinct depths within the vertebral body and different meeting points with angle $\theta$. Figure \ref{fig:Param} shows the X-ray overlay of these scenarios. 

\begin{figure}[t] 
    \centering 
    \includegraphics[width=1\linewidth]{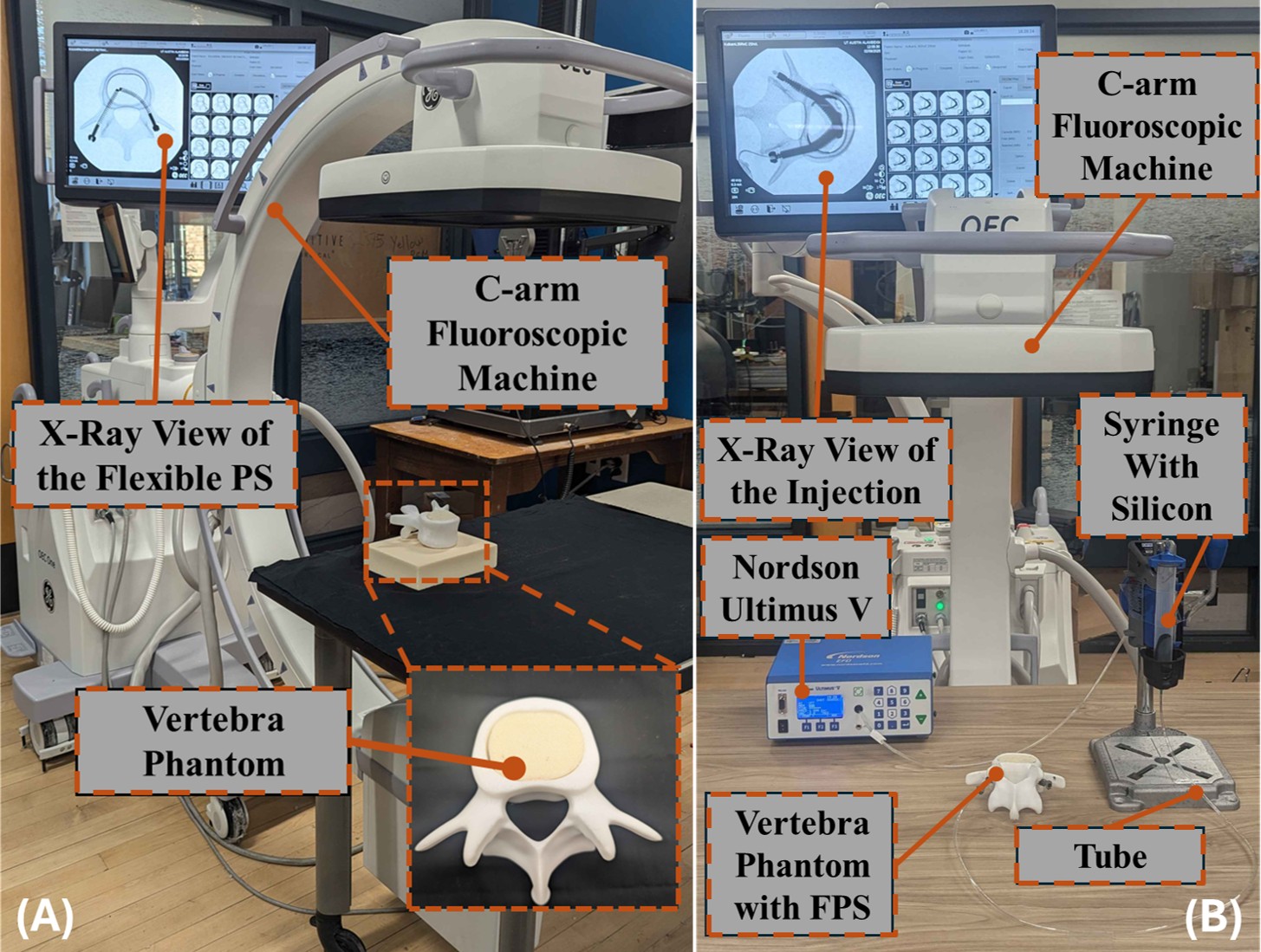}
    \caption{(A) Experimental set-up used for taking X-ray shots and visualizing the FPS insertion process in the drilled tunnels. (B) Experimental set-up used for silicon injection for the simulated AB-SF procedure. }
    \label{fig:XRay}
\end{figure}

In \textit{Scenario 1 (S1)}, to mimic a realistic SF surgical procedure, we assumed a surgeon, using the aforementioned planning procedures, picked a scenario matching the combination of \textit{Straight-Curve} AB-SF trajectory shown in Fig. \ref{fig:Param}. For this configuration, we assumed the surgeon picked the Curved side trajectory with a $L_{OT}$ of 28.5 mm and a $L_{IT}$ of 42.5 mm with a $r$ of 25 mm. For the Straight side, the surgeon picked an $L_{OT}$ of 49.4 mm. To ensure a connection between the two FPSs, a meeting point angle ($\theta$) of approximately 110 degrees was chosen. These dimensions are shown on Fig. \ref{fig:Param}. After the surgeon selected the optimal trajectory, the NiTi tube for the CT-SDR was heat treated to the 25 mm radius and the CT-SDR was assembled together for the procedure as shown in the Fig. \ref{fig:Experiment}. For this procedure, a patient-specific vertebral phantom was 3D printed from PLA material (Raise3D). As shown in Fig. \ref{fig:Experiment}, only the outer shell of the vertebral phantom was printed to mimic hard cortical bone of the vertebra. During the printing process, a 8 mm hollow corridor was incorporated in pedicle region of the vertebral phantom, replicating the straight part of the drilling trajectory. The inner vertebral body was left empty so that it could be filled with Sawbone material (Pacific Research Laboratories, USA) -- a synthetic bone substitute. For this study, PCF 5 Sawbone was used to mimic extremely osteoporotic trabecular bone \cite{ccetin2021experimental}. Of note,   for this proof-of-concept study, the vertebral phantom used was scaled to be 1.5 times the size of an L3 vertebra. The dimensions for the vertebral phantom are shown in Fig. \ref{fig:Experiment}.

\begin{figure*}[t] 
    \centering 
    \includegraphics[width=1\linewidth]{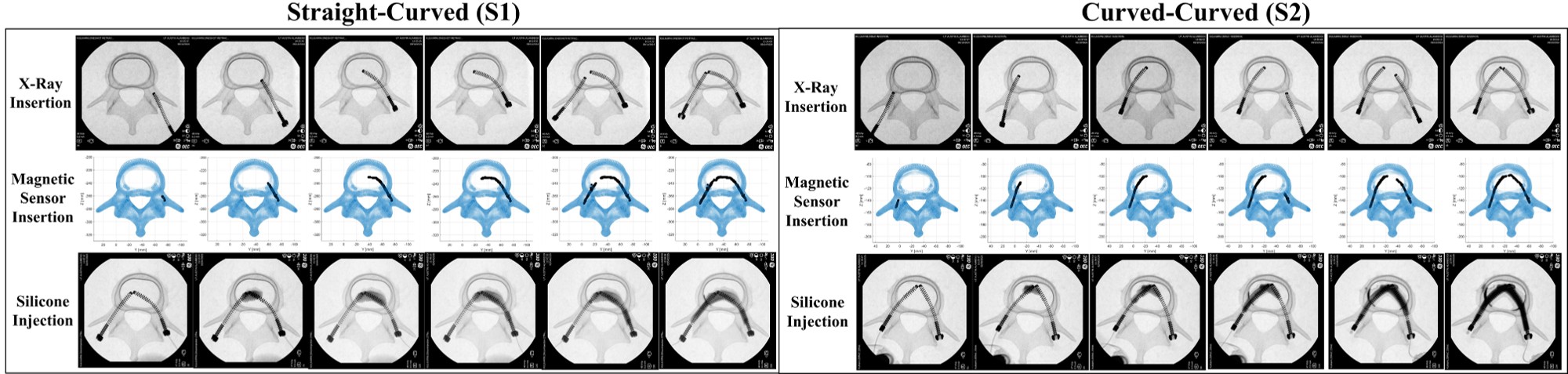}
    \caption{First row of the figure shows the snapshots of inserting the FPSs inside the drilled tunnels in S1 (Straight-Curved) and S2 (Curved-Curved) scenarios. These figures  clearly show morphability of the FPSs inside the drilled tunnels. Second row of the figure also demonstrates the results of ICP registration algorithms and the measured trajectory of the inserted FPSs by the magnetic tracker in both scenarios of AB-SF. Last row of figure also illustrates the feasibility of the AB-SF concept by injecting silicon inside the inserted screws for both S1 and S2 scenarios.}
    \label{fig:Results}
\end{figure*}

As shown in Fig. \ref{fig:Experiment}, after the vertebral body was securely fixed within the surgical workspace, the CT-SDR system was activated to create the aforementioned curved trajectory dimensions. The CT-SDR was placed into admittance control mode  to allow the surgeon to freely move and align the robot with the starting point of the pedicle region of the vertebral body. Once the robot was properly aligned, the CT-SDR's drill tip was accelerated to an approximate speed of 6000 rpm. The autonomous drilling strategy was activated and the $L_{OT}$ of 28.5 mm was drilled first at a speed of 2 mm/s. Once the robot moved the prescribed $L_{OT}$ distance, the stationary control strategy was activated and curved part was drilled by inserting the NiTi tube of the CT-SDR at the same speed. Once the appropriate distance had been reached the drill tip rpm was reduced down to 1000 rpm and the drill was reversed at the same 2 mm/s speed. The drill was then realigned on the other pedicle and only the autonomous drilling strategy was activated to create a straight trajectory of 49.4 mm with aforementioned speed and rpm values. 
As illustrated in Fig. \ref{fig:Param}, in \textit{scenario 2 (S2)}, we created two symmetric J-shape trajectories representing a combination of asymmetric \textit{Curved-Curved} AB-SF trajectory in a deeper part of the vertebral body (as compared with S1). The described drilling process for creating a curved trajectory in S1, was repeated  twice to create this connecting drilling trajectory shown in Fig. \ref{fig:Param}. This configuration had a $L_{OT}$ of 36.6 mm and a $L_{IT}$ of 30.9 mm with a $r$ of 35 mm. This same trajectory was used on both sides of the pedicle. The surgeon picked a meeting angle ($\theta$) of approximately 89 degrees. These dimensions are shown in Fig. \ref{fig:Param}.
After all trajectories were drilled, a 5 DoF magnetic sensor (Norther Digital Inc.) was embedded and fixated through the hollow middle of a FPS. For this experimental procedure, guided by the geometry and dimensions of an L3 vertebra \cite{Zindrick1986ABS}, a FPS with a rigid section of 18 mm ($L_R$) was designed to correspond with the length of the pedicle region. The flexible portion of the FPS measures 54.4 mm ($L_F$). The FPS has an outer diameter (OD) of 7 mm to align with the standard size of pedicle screws used in surgery, and an inner diameter (ID) of 4 mm. The threads are uniformly spaced with a pitch of 2.5 mm. The FPS was designed using the Direct Metal Laser Sintering (DMLS) process with Ti-6Al-4V grannulates using a Renishaw AM 250 Laser Melting System (Leuven, Belgium). Ti-6Al-4V was choosen based on ASTM F3001 standards \cite{ASTMF3001}. The dimensions of the FPS utilized can be seen in Fig. \ref{fig:Experiment}. The 5 DoF magnetic sensor was used to quantitatively analyze the performance of the drilling procedure by measuring the radius of curvature of the implanted FPS within the drilled trajectories. After the 5 DoF magnetic sensor was safely and securely fixated within the FPS and before the radius of curvature were collected, an Iterative Closest Points (ICP) registration algorithm \cite{ICP_ZHOU202263} was performed to align the vertebra's 3D model reference frame with the magnetic tracker frame. The insertion and retraction measurement is then performed three times for both S1 and S2 scenarios shown in Fig. \ref{fig:Param}. The formation of the AB-SF structure was further observed independent of the magnetic tracking system under a C-arm X-ray machine (OEC One CFD, GE Healthcare)  shown in Fig. \ref{fig:XRay}-A. The results of ICP registration algorithm and calculated errors can be seen in Fig. \ref{fig:Results} and Table \ref{table:1}. Also, Fig. \ref{fig:Results} shows the snapshots of inserting the FPSs inside the drilled tunnels in S1 and S2 scenarios, clearly showing morphability of the FPSs inside the drilled tunnels.  

After the magnetic tracking data was collected, we performed a feasibility study to validate the augmented bridge concept. Without loss of generality, to simulate the augmentation process and  assess the feasibility of material injection, a mixture of magnetic powder with Ecoflex 00-10 silicone (Smooth-On, Inc.) was used as a substitute to bone cement. This silicone has a viscosity of 14,000 cP, as noted in \cite{SmoothOnEcoflex}. Specifically, the mixture consisted of approximately 10 g of Ecoflex 00-10 silicone Part A, 10 g of Ecoflex 00-10 silicone Part B, 1 g of Dowsil OS-10 (Dow, USA), and 10 g of magnetic powder (Neo Magnequench, Singapore). This composition was selected to achieve a smoother consistency and ensure a sufficient concentration of magnetic particles for clear visualization in X-ray imaging.  As shown in Fig. \ref{fig:XRay}-B, after preparing the silicone mixture and loading it into a syringe, a silicone tube (2 mm outer diameter) was inserted through  the internal space of FPS until it reached the meeting point of two FPSs. A pneumatic dispenser (Nordson Ultimus V, Nordson EFD, U.S.A.) was then used to inject the mixture through the tube and along the fixated FPS path under a pressure of 4 Bar. As the silicone mixture was injected, X-ray images were taken using the C-arm x-ray machine to ensure the space is completely filled with the injected material and help with retracting the injection tube. Figure \ref{fig:Results} shows the snapshots of silicone injection into the FPSs for both S1 and S2 scenarios.

\section{Discussion}
The X-ray images in Fig. \ref{fig:Results} illustrate the FPS behavior at various stages of the insertion process inside the drilled J-shape trajectories for both S1 and S2 scenarios. These snapshots clearly demonstrate the performance of the steerable drilling system to create two distinct AB-SF trajectories in different depths of the vertebral body that can meet at the planned points inside the vertebra. Also, they show the FPS's capability to safely follow both AB-SF configurations and then connect with each other at a planned location. This demonstration further validates the system's ability to conduct FPS fixation on either sides of the vertebral body or within the middle of the vertebral body. For the expected 25 mm radius of curvature ($r$) associated with the S1 scenario, we measured an average of 26.84 mm radius of curvature, resulting in a low error of 7.4\%. Furthermore, the test had a low standard deviation of 0.83 mm, further validating the system's capability to accurately create and follow a tight radius of curvature. For the S2 scenario, we measured an average radius of curvature of 38.36 mm, resulting in a low error of 9.6\% (compared to an ideal 35 mm radius of curvature) with a 4.05 mm standard deviation. Table \ref{table:1} summarizes the results of this quantitative analysis.

Figure \ref{fig:Results} also illustrates all the insertion trajectories gathered for both configurations using the magnetic sensor data inside of the vertebral phantom. This visualization and quantitative assessment is vital in ensuring the CT-SDR is properly creating J-shape tunnels and  the FPSs are following these trajectories and fixating the vertebra without deviating from the planned trajectory. The visualization also provides vital information on whether both components successfully created an augmented bridge formation. Furthermore, these plots are crucial in identifying the changeover location between the straight and curved trajectory. This is then used to calculate the radius of curvature using a best fit. Of note,  to evaluate the accuracy of the used ICP algorithm  \cite{ICP_ZHOU202263}, the root mean square error (RMSE) for this  registration method was calculated. For both configurations, we saw a combined low RMSE error of 0.98 mm with the individual RMSE for each model shown in Table \ref{table:1}. This low error validates the accuracy of the performed procedure.

The X-ray images of the silicone injection process in Fig. 5 provide critical validation of the AB-SF concept, demonstrating the feasibility of forming a continuous, reinforced bridge between the two FPSs. These images clearly capture the material's flow behavior as it disperses through the FPS and around its structure, effectively replicating the envisioned augmentation process with a bone cement. The ability to achieve this bridge formation in multiple trajectory configurations reinforces the robustness and adaptability of the AB-SF technique. Also, the bone cement can create a continuous bridge between the created J-shape tunnels that biomechanically can distribute the load exerted on the vertebrae and minimize the risk of screw loosening and pullout \cite{Sharma2024TBME}. Overall, the performed experiments and results  demonstrate the potentials of the utilized steerable drilling system and FPSs in performing an AB-SF procedure. As mentioned, this procedure has the potential to address the pullout and loosening failures of existing RPSs by reinforcing a spinal fixation procedure. 

\begin{table}
\centering
\caption{Experimental Results across trials completed for each Augmented Bridge trajectory}
\label{table:1}
\setlength\tabcolsep{0pt} 
\begin{tabular*}{1\columnwidth}{@{\extracolsep{\fill}} l c c c c}
\Xhline{1.25pt}
Trajectory & \makecell{Ideal $r$ \\ (mm)} & \makecell{Actual $r$ \\ (mm)} & Error & \makecell{ICP RMSE \\ (mm)} \\
\Xhline{1.25pt}
Straight-Curved & 25 & 26.84 $\pm$ 0.83 & 7.4\% & 1\\
\Xhline{0.25pt}
\makecell{Curved-Curved: Left\\Hand Side} & 35 & 38.36 $\pm$ 4.05 & 9.6\% & 0.95\\
\Xhline{0.25pt}
\Xhline{1.25pt}
\end{tabular*}
\smallskip
\scriptsize
\end{table}

\section{Conclusion and Future Work}
In this paper, leveraging the CT-SDR robotic system and FPS implants, we  proposed a novel spinal fixation procedure for an osteoporotic vertebra. For the first time to our knowledge, we  (i) introduced the novel concept of AB-SF to connect two FPSs using an injected material (e.g., bone cement) and minimize the risk of screw pullout and loosening;  (ii)  developed a comprehensive trajectory planning framework for AB-SF procedure; and (iii)  successfully assessed the proposed novel fixation concept by executing two different Straight-Curved and Curved-Curved AB-SF scenarios. In the future, we will focus on performing a more realistic AB-SF procedure using a real bone cement (e.g., PMMA or calcium phosphate \cite{Shea2014DesignsAT}). We will also  repeat our experiments on animal bones and cadaveric specimens.
 
\bibliographystyle{IEEEtran}
\bibliography{root}

\end{document}